\documentclass[10pt,twocolumn,letterpaper]{article}

\usepackage{iccv}
\usepackage{times}
\usepackage{epsfig}
\usepackage{graphicx}
\usepackage{amsmath}
\usepackage{amssymb}
\usepackage{color}
\usepackage{multirow}
\usepackage{algorithm}
\usepackage{algorithmic}

\usepackage{subcaption}
\usepackage{float}

\usepackage[pagebackref=true,breaklinks=true,letterpaper=true,colorlinks,bookmarks=false]{hyperref}

\iccvfinalcopy 

\def\iccvPaperID{3616} 

\ificcvfinal\fi
\begin{document}

\title{Improving Object Detection with Inverted Attention}

\author{Zeyi Huang\\
Carnegie Mellon University\\
{\tt\small zeyih@andrew.cmu.edu}
\and
Wei Ke\\
Carnegie Mellon University\\
{\tt\small weik@andrew.cmu.edu}
\and
Dong Huang\\
Carnegie Mellon University\\
{\tt\small donghuang@cmu.edu}
}

\maketitle

\begin{abstract}
 Improving object detectors against occlusion, blur and noise is a critical step to deploy detectors in real applications. Since it is not possible to exhaust all image defects through data collection, many researchers seek to generate hard samples in training. The generated hard samples are either images or feature maps with coarse patches dropped out in the spatial dimensions. Significant overheads are required in training the extra hard samples and/or estimating drop-out patches using extra network branches. In this paper, we improve object detectors using a highly efficient and fine-grain mechanism called Inverted Attention (IA). Different from the original detector network that only focuses on the dominant part of objects, the detector network with IA iteratively inverts attention on feature maps and puts more attention on complementary object parts, feature channels and even context. Our approach (1) operates along both the spatial and channels dimensions of the feature maps; (2) requires no extra training on hard samples, no extra network parameters for attention estimation, and no testing overheads. Experiments show that our approach consistently improved both two-stage and single-stage detectors on benchmark databases.
\end{abstract}


\section{Introduction}

\begin{figure}[t]
\centering
 \centerline{\textbf{Attention heat-maps without Inverted Attention (IA)}}
    \begin{subfigure}[b]{0.18\linewidth}
      \includegraphics[width=\linewidth]{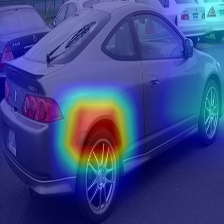}
    \end{subfigure}
     \begin{subfigure}[b]{0.18\linewidth}
       \includegraphics[width=\linewidth]{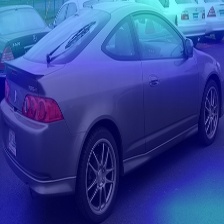}
    \end{subfigure}
     \begin{subfigure}[b]{0.18\linewidth}
       \includegraphics[width=\linewidth]{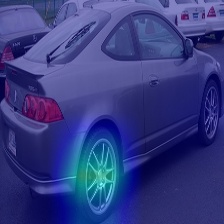}
     \end{subfigure}
     \begin{subfigure}[b]{0.18\linewidth}
       \includegraphics[width=\linewidth]{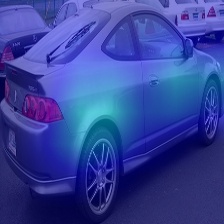}   
      \end{subfigure}
       \hspace{0.05cm}
        \begin{subfigure}[b]{0.18\linewidth}
      \includegraphics[width=\linewidth]{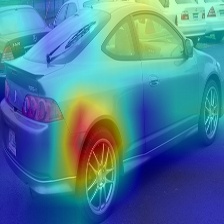}
       \end{subfigure}
    \\
    \centerline{\textbf{Inverted attention heat-maps during IA training}}
    
    \begin{subfigure}[b]{0.18\linewidth}
      \includegraphics[width=\linewidth]{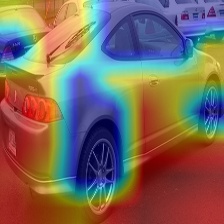}
    \end{subfigure}
     \begin{subfigure}[b]{0.18\linewidth}
       \includegraphics[width=\linewidth]{fig/channel1/gt_314_5_inverted_new.jpg}
    \end{subfigure}
     \begin{subfigure}[b]{0.18\linewidth}
       \includegraphics[width=\linewidth]{fig/channel1/gt_314_5_inverted_new.jpg}
     \end{subfigure}
     \begin{subfigure}[b]{0.18\linewidth}
       \includegraphics[width=\linewidth]{fig/channel1/gt_314_5_inverted_new.jpg}   
      \end{subfigure}
       \hspace{0.05cm}
        \begin{subfigure}[b]{0.18\linewidth}
      \includegraphics[width=\linewidth]{fig/channel1/gt_314_5_inverted_new.jpg}
       \end{subfigure}
       \\
    \centerline{\textbf{Attention heat-maps trained with IA}}
    \begin{subfigure}[b]{0.18\linewidth}
      \includegraphics[width=\linewidth]{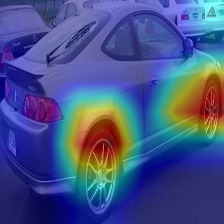}
     \end{subfigure}
     \begin{subfigure}[b]{0.18\linewidth}
       \includegraphics[width=\linewidth]{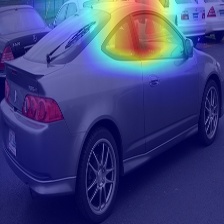}
    \end{subfigure}
     \begin{subfigure}[b]{0.18\linewidth}
       \includegraphics[width=\linewidth]{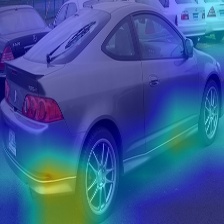}
     \end{subfigure}
     \begin{subfigure}[b]{0.18\linewidth}
       \includegraphics[width=\linewidth]{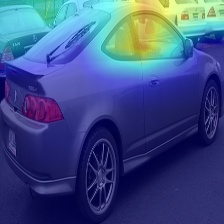}
       \end{subfigure}
       \hspace{0.05cm}
     \begin{subfigure}[b]{0.18\linewidth}
     \includegraphics[width=\linewidth]{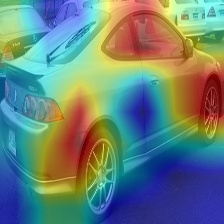}
     \end{subfigure}
     \leftline{\hspace{0.03 cm} $|$\hspace{0.32 cm}-Top 4 feature channels affected by IA-\hspace{0.32 cm}$|$\hspace{0.01 cm} $|$ -Overall- $|$}
\caption{Attention heat-maps visualized at the ROI feature map ($[7\times 7\times 512]$) of VGG16 Fast-RCNN. Each $[7\times 7]$ attention heatmap is linearly interpolated and superposed on the $[224\times 224]$ object ROI image patch. The first 4 columns are respectively the attention heat-maps at the top 4 channels affected by IA. The last column, ``Overall",  denotes the overall attention summed over all $512$ channels. (This figure is best viewed in color)}
\label{fig:channel_IA}
\vspace{-0.2cm}
\end{figure}

Improving object detectors against image defects such as occlusion, blur and noise is a critical step to deploy detectors in real applications. Recent efforts by computer vision community have collected extensively training data on different scenes, object categories, shapes and appearance. However, it is yet not possible to exhaust all image defects captured under camera shake, dust, fade lighting and tough weather conditions. Moreover, deep learning approaches are highly biased by data distribution, while it is very difficult to collect data with uniform combinations of object features and defects.   

Besides waiting for more data collection and annotation, a fundamental way to improve detectors is to improve the training approach. Four main attempts have been recently explored by the computer vision community: (1) Selecting subsets of training samples by hard example mining~\cite{shrivastava2016training}. This approach does not generalize well to unseen images defects. (2) Penalizing the occluded bounding boxes by occlusion-aware losses ~\cite{wang2018repulsion,zhang2018occlusion}. These approaches do not generalize well to unseen occlusion. (3) Synthesizing image defects by hard example generation~\cite{singh2017hide,wang2017fast,zhang2018occluded, zhong2017random}. These approaches typically drop out big patches in spatial dimension and the new samples to be trained increase exponentially. (4) Highlighting desired fine-grain features by attention estimation~\cite{fu2018dual,hu2018squeeze,selvaraju2017grad,woo2018cbam,zhou2016learning}. Estimating the attention masks usually require extra networks and therefore increase the training overheads.

\begin{figure*}[t]
  \centering
  \centerline{Overall attention heat-maps trained \textbf{without} IA}
   \begin{subfigure}[b]{0.12\linewidth}
      \includegraphics[width=\linewidth]{fig/before/gt_314_5.jpg}
       \end{subfigure}
      \begin{subfigure}[b]{0.12\linewidth}
       \includegraphics[width=\linewidth]{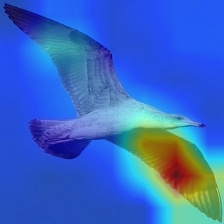}
     \end{subfigure}
     \begin{subfigure}[b]{0.12\linewidth}
      \includegraphics[width=\linewidth]{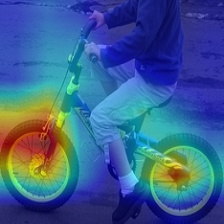}
       \end{subfigure}
      \begin{subfigure}[b]{0.12\linewidth}
       \includegraphics[width=\linewidth]{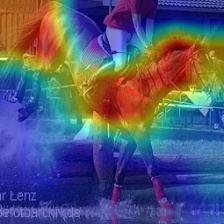}
     \end{subfigure}
     \begin{subfigure}[b]{0.12\linewidth}
      \includegraphics[width=\linewidth]{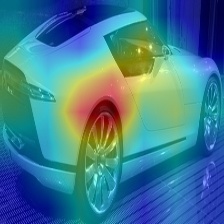}
       \end{subfigure}
      \begin{subfigure}[b]{0.12\linewidth}
       \includegraphics[width=\linewidth]{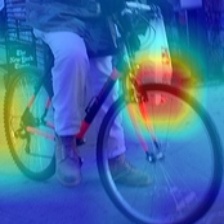}
     \end{subfigure}
     \begin{subfigure}[b]{0.12\linewidth}
      \includegraphics[width=\linewidth]{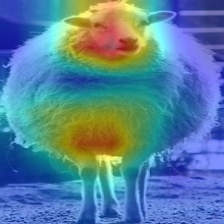}
       \end{subfigure}
      \begin{subfigure}[b]{0.12\linewidth}
       \includegraphics[width=\linewidth]{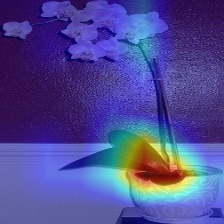}
     \end{subfigure}
    \\
    \centerline{Overall attention heat-maps trained \textbf{with} IA}
    \begin{subfigure}[b]{0.12\linewidth}
     \includegraphics[width=\linewidth]{fig/after/gt_314_5.jpg}
    \end{subfigure}
     \begin{subfigure}[b]{0.12\linewidth}
       \includegraphics[width=\linewidth]{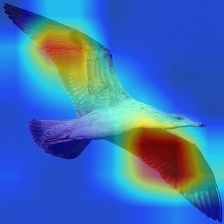}       
      \end{subfigure}
      \begin{subfigure}[b]{0.12\linewidth}
      \includegraphics[width=\linewidth]{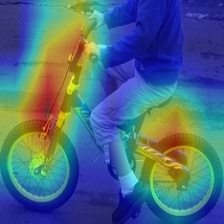}
       \end{subfigure}
      \begin{subfigure}[b]{0.12\linewidth}
       \includegraphics[width=\linewidth]{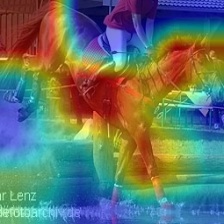}
     \end{subfigure}
     \begin{subfigure}[b]{0.12\linewidth}
      \includegraphics[width=\linewidth]{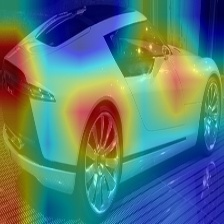}
       \end{subfigure}
      \begin{subfigure}[b]{0.12\linewidth}
       \includegraphics[width=\linewidth]{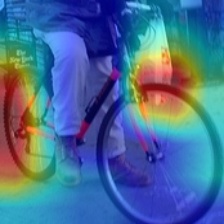}
     \end{subfigure}
     \begin{subfigure}[b]{0.12\linewidth}
      \includegraphics[width=\linewidth]{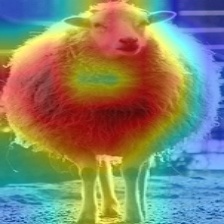}
       \end{subfigure}
      \begin{subfigure}[b]{0.12\linewidth}
       \includegraphics[width=\linewidth]{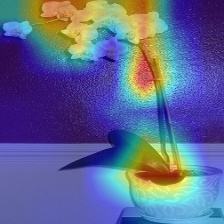}
     \end{subfigure}
     \caption{More examples of overall attention heat-maps trained without Inverted Attention (\textbf{Upper row}), and trained with Inverted Attention (\textbf{Lower row}).}
     \label{fig:overall_IA}
     \vspace{-0.2cm}
\end{figure*}

In this paper, we improve object detector using a highly efficient and fine-grain mechanism called Inverted Attention (IA). IA is implemented as a simple module added to the standard back-propagation operation. In every training iteration, IA computes the gradient of the feature maps produced at the feature extraction network (also called the backbone network) using object classification scores, and iteratively invert the attention of the network. Different from the original detector network that only focuses on the small parts of objects, the neural network with IA puts more attention on complementary spatial parts of the original network, feature channels and even the context. The IA module only changes the network weights in training and does not change any computation in inference.

Fig \ref{fig:channel_IA} (the upper row) visualizes the attention in a standard Fast-RCNN detector. The attention is visualized as heatmaps superposed on the object. The red pixels of the heat-maps denote high attention, while the blue pixels denote low attention. During our IA training, the original attention heat-maps are inverted as Fig \ref{fig:channel_IA} (the middle row) and proceed to the next iteration. After the IA training finishes, the network produces new attention heat-maps, see Fig \ref{fig:channel_IA} (the lower row). Observe that, the detector network trained with IA focuses on more comprehensive features of the objects, making it more robust to the potential defects of the individual pixels. Fig \ref{fig:overall_IA} shows more examples of overall attention produced without IA training (the upper row) and with IA training (the lower row). Comparing to the existing approaches, our approach (1) operates along both the spatial and channels dimensions of the feature maps; (2) requires no extra training on hard samples, no extra network parameters for attention estimation, and no testing overheads. We evaluated IA on both the two-stage and single-stage detectors on benchmark databases, and produced significant and consistent improvement.


\begin{figure*}
\centering
\includegraphics[width=0.95\textwidth]{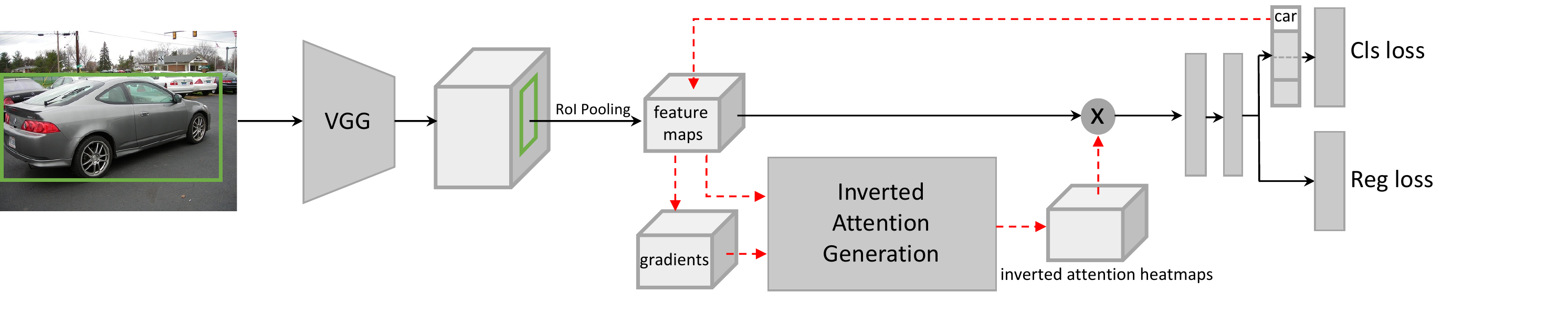}
\caption{Architecture of Inverted Attention Network (IAN) based on a RNN object detection network. Light gray blocks denote tensors. Dark gray blocks denote operations. The data flows along the red dashed arrows are only needed in training. The inference only require the flows along the black arrows.}
\label{fig:network}
\end{figure*}

\section{Related Work}
There are three kind of approaches to improve object detectors against noisy data. 

\subsection{Occlusion-Aware Loss}
\cite{zhang2018occlusion} proposed a aggregation loss for R-CNN based person detectors. This loss enforced proposals of the same object to be close. Instead of using a single RoI pooling layer for a person, the authors used a five-part pooling unit for each person to deal with occlusions. \cite{wang2018repulsion} proposed a new bounding box regression loss, termed repulsion loss. This loss encourages the attraction by target, and the repulsion by other surrounding objects. The repulsion term prevents the proposal from shifting to surrounding objects thus leading to more crowd-robust localization. \cite{zhou2018bi} proposed two regression branches to improve pedestrian detection performance. The first branch is used to regress full body regions. The second branch is used to regress visible body regions. 

\subsection{Hard Sample Generation}
\textbf{Image based generation:}~\cite{zhong2017random} randomly occludes several rectangular patches of images during training.~\cite{huang2018adversarially} occludes rectangular patches of image guided by the loss of the person Re-Identification task. \cite{singh2017hide} improves weakly-supervised object localization by randomly occluding patches in training images. \cite{wang2017fast} learns an adversarial network that generates examples with occlusions and deformations. The goal of the adversary is to generate examples that are difficult for the object detector to classify. 

\textbf{Feature based generation:}~\cite{hou2019weighted} selects weighted channel to dropout for the regularization of CNNs.~\cite{wang2017fast} uses predicted occlusion spatial mask to dropout for generating hard positive samples.~\cite{zhang2018occluded} re-weights feature maps according to three kinds of channel-wise attention mechanism.~\cite{song2018vital} adaptively dropout input features to obtain various appearance changes using random generated masks in video tracking.
Our IA approach conducts both channel-wise and spatial-wise dropout on features.

\subsection{Attention Estimation}
Recent methods incorporate attention mechanism to improve the performance of CNNs. \cite{fu2018dual,hu2018squeeze,woo2018cbam} integrate channel-wise or spatial-wise attention network branches to the feed-forward convolutional neural networks. The estimated attention maps are multiplied to the original feature map for various CNN tasks. All these methods introduce extra network branches to estimate the attention maps. \cite{zagoruyko2016paying} improves the performance of a student CNN network by transferring the attention maps of a powerful teacher network. Deconvnet~\cite{zeiler2014visualizing} and guided-backpropagation~\cite{springenberg2014striving} improve gradient-based attention using different backpropagation strategy. CAM~\cite{zhou2016learning} converts the linear classification layer into a convolutional layer to produce attention maps for each class. Grad-CAM~\cite{selvaraju2017grad} improve CAM and is applicable to a wide variety of CNN model. 

In our method, we compute inverted attention guided by gradient-based attention and Grad-CAM. Our network does not require extra network parameters or teacher networks.


\section{Inverted Attention for Object Detection}

\begin{figure}[t]
\centering
\includegraphics[width=1\linewidth]{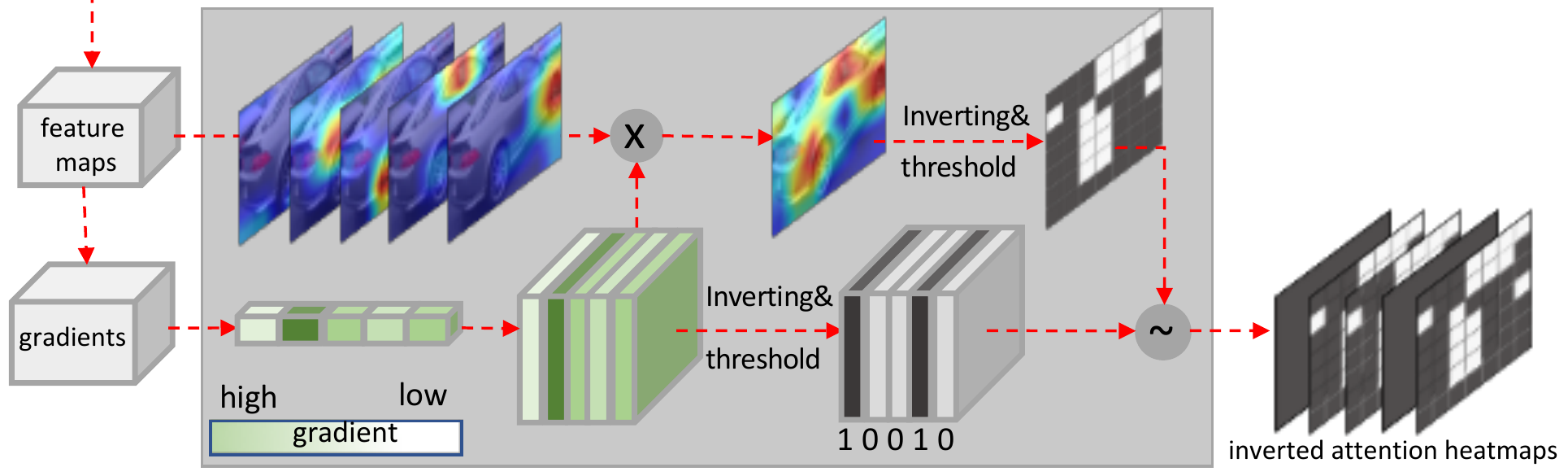}
\caption{Detailed architecture of the Inverted Attention Generation Module in Fig.~\ref{fig:network}. This module consists of only simple operations such as pooling, threshold and element-wise product. No extra parameter is needed to learn. This module only operates during training, not in testing.}
\label{fig:IAG}
\end{figure}

We take the R-CNN \cite{girshick2015fast} based framework to illustrate our Inverted Attention Network (IAN) (Fig.~\ref{fig:network}). An Inverted Attention Generation Module (Fig.~\ref{fig:IAG}) is added to the R-CNN detection network, and operates on the ROI featrue maps. Note that this module consists of only few simple operations such as pooling, threshold and element-wise product. No extra parameter is needed to learn. In the rest of the paper, we show how this simple change can effectively improve the original network to overcome image defects.  


\subsection{Inverted Attention Generation Module}

The attention mechanism identifies the most representative receptive field of an object. Highlighting the object with attention heatmaps encourages discrimination between object classes, meanwhile, decreases the diversity of features within the same class. However, the diversity of features is the key to generalize an object detector to unseen object instances and image defects. The proposed Inverted Attention training approach aims to alleviate the conflict and find the optimal trade-off between discrimination and diversity. 

\begin{figure*}[h]
  \centering
  \begin{subfigure}[b]{0.12\linewidth}
      \includegraphics[width=\linewidth]{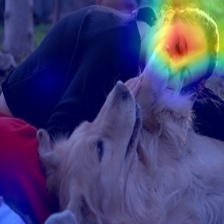}
     \end{subfigure}
     \begin{subfigure}[b]{0.12\linewidth}
      \includegraphics[width=\linewidth]{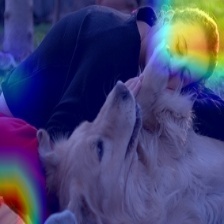}
     \end{subfigure}
     \begin{subfigure}[b]{0.12\linewidth}
      \includegraphics[width=\linewidth]{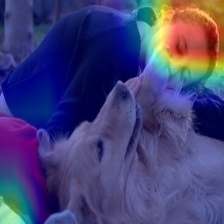}
      \end{subfigure}
     \begin{subfigure}[b]{0.12\linewidth}
      \includegraphics[width=\linewidth]{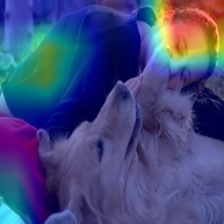}       
     \end{subfigure}
    \begin{subfigure}[b]{0.12\linewidth}
      \includegraphics[width=\linewidth]{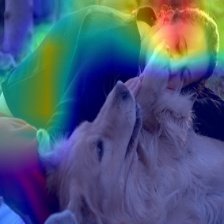}
      \end{subfigure}
     \begin{subfigure}[b]{0.12\linewidth}
      \includegraphics[width=\linewidth]{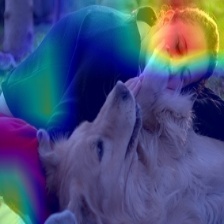}
     \end{subfigure}
     \begin{subfigure}[b]{0.12\linewidth}
      \includegraphics[width=\linewidth]{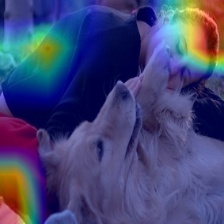}
     \end{subfigure}
     \begin{subfigure}[b]{0.12\linewidth}
      \includegraphics[width=\linewidth]{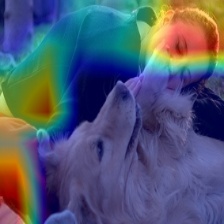}       
      \end{subfigure}
    \\
    \begin{subfigure}[b]{0.12\linewidth}
      \includegraphics[width=\linewidth]{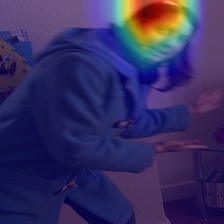}
      \caption{Epoch 1}
     \end{subfigure}
     \begin{subfigure}[b]{0.12\linewidth}
      \includegraphics[width=\linewidth]{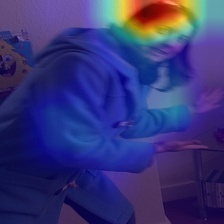}
      \caption{Epoch 2}
    \end{subfigure}
     \begin{subfigure}[b]{0.12\linewidth}
      \includegraphics[width=\linewidth]{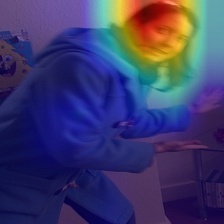}
      \caption{Epoch 3}
     \end{subfigure}
     \begin{subfigure}[b]{0.12\linewidth}
      \includegraphics[width=\linewidth]{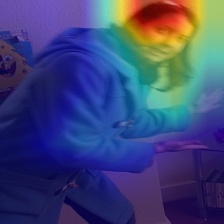}       
      \caption{Epoch 4}
     \end{subfigure}
    \begin{subfigure}[b]{0.12\linewidth}
      \includegraphics[width=\linewidth]{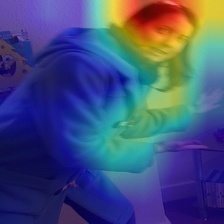}
      \caption{Epoch 5}
     \end{subfigure}
     \begin{subfigure}[b]{0.12\linewidth}
      \includegraphics[width=\linewidth]{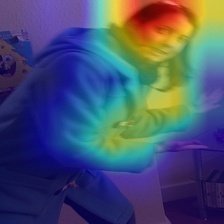}
      \caption{Epoch 6}
    \end{subfigure}
     \begin{subfigure}[b]{0.12\linewidth}
      \includegraphics[width=\linewidth]{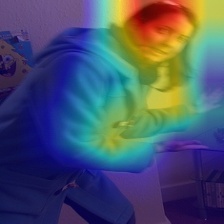}
      \caption{Epoch 7}
     \end{subfigure}
     \begin{subfigure}[b]{0.12\linewidth}
      \includegraphics[width=\linewidth]{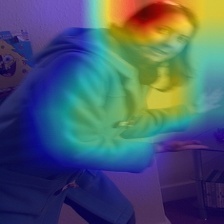}       
      \caption{Epoch 8}
     \end{subfigure}
     \caption{Network attention evolves at different epochs during the IA training. Each sub-image shows the 2D attention heat-maps superposed on object ROI. The attention maps are visualized at the ROI feature map ($[7\times 7\times 512]$) of the VGG16 backbone network in Fast-RCNN.}
     \label{fig:IAepoch}
\end{figure*}

As shown in Fig.~\ref{fig:IAG}, the Inverted Attention Generation Module consists of two simple operations: (1) Gradient-Guided Attention Generation: computing the the gradients at feature maps, by back-warding only the classification score on the ground-truth category, (2) Attention Inversion: reversing element values of the attention tensor to produce IA heat-maps. 

\textbf{Gradient-Guided Attention Generation:} In training phase of the convolutional neural networks, gradients of feature maps in the back-propagation operation encode how sensitive the output prediction is with respect to changes at the same location of the feature maps. If small changes at an element of feature maps have a strong effect on the network outputs, then the network will be updated to pay more attention on that feature element. With this principle, we use the gradient to generate attention map in our approach (see details in Fig.\ref{fig:IAG}). 

Denote the gradient tensor as $G$ and feature tensor as $F$. Both of them are of size $H \times W \times C$, where $H, W, C$ are the height, width and channel number of $G$ and $F$, respectively. A global pooling is applied on the gradient tensor to produce a weight vector $W$ with size $C \times 1$. We compute a gradient guided attention map following \cite{selvaraju2017grad}, 
\begin{equation}
    M = \sum_{i}^{C} w_{i} * F^{(i)},
\label{eq:mask}
\end{equation}
where $w_i$ is the $i$-th element of $W$, and $F^{(i)}$ is the $i$-th channel map of $F$. As shown in Fig.~\ref{fig:IAG}, the high values in gradient correspond to the receptive field of and trunk in the car sample, while small values correspond to the receptive field of door of the car and background.

\textbf{Attention Inversion:} In the standard training process, the gradient descent algorithm forces the attention map to converge to a few most sensitive parts of objects, while ignores the other less sensitive parts of objects. The IA training conducts iterative inverting of the original attention tensor as the \textit{Inverted Attention} tensor, which forces the network to detect object based on their less sensitive parts. Specifically, we generate a spatial-wise inverted attention map and a channel-wise inverted attention vector, and then combine them to produce the final attention maps.

The spatial-wise inverted attention map $A^{s} = \{a^{s}_{i}\}$ is computed as
\begin{equation}
a^{s}_{i} = \begin{cases}
0 & \text{ if } m_i> T_{s} \\ 
1 & \text{ else } 
\end{cases} ,
\label{eq:sa}
\end{equation}
where $a^{s}_{i}$ and $m_{i}$ are the elements of $A^{s}$ and $M$ at the $i$-th pixel, respectively. $T_{s}$ is the threshold for spatial-wise attention map. From Eq.~\ref{eq:sa}, spatial-wise inverted attention map pays more attention to the area of the sample with small gradient value.

Observe that the weight vector $W$ serves as a sensitivity measure for channels of feature maps. A threshold $T_{c}$ is used to compute the channel-wise inverted attention vector $A^{c}=\{ a^{c}_{j} \}$,
\begin{equation}
a^{c}_{j} = \begin{cases}
0 & \text{ if } w_j > T_{c} \\ 
1 & \text{ else } 
\end{cases} .
\label{eq:ca}
\end{equation}
The final Inverted attention map $A = \{ a_{i,j}\}$ is computed as
\begin{equation}
a_{i,j} = \begin{cases}
a^{s}_{i} & \text{ if } a^{c}_{j} = 0 \\ 
1 & \text{ else } 
\end{cases} .
\end{equation}

 Fig.~\ref{fig:IAepoch}) illustrates how network attention evolves at different epochs during the IA training. The IA training iteratively guides the neural network to extract features on the whole object sample.

\subsection{Inverted Attention Network}

As shown in Fig.~\ref{fig:network}, the Inverted Attention Network (IAN) is basically built by adding Inverted Attention Generation Module (Fig.~\ref{fig:IAG}) to the R-CNN based detection network, and operates on the ROI feature maps.

Given an input image, the backbone of the R-CNN framework, \textit{i.e.}, VGG or ResNet, takes the whole image as an input and produces feature maps. The region proposals are generated from these feature maps by region proposal network (RPN) or pre-computed region proposal candidates. Then the RoI-pooling layer generates a fixed size feature maps for each object proposal. These feature maps after RoI pooling then go through fully connected layers for object classification and bounding box regression. 

The R-CNN can be trained end-to-end by optimizing the following two loss functions:
\begin{equation}
    L_{rpn} = L_{cross-entropy} + L_{rpn\_reg},
\label{equ:l1}
\end{equation}
\begin{equation}
    L_{rcnn} = L_{softmax} + L_{rcnn\_reg},
\label{eq:cls}
\end{equation}
where~$L_{cross-entropy}$ and ~$L_{rpn\_reg}$ are the cross-entropy loss and L1 loss for RPN network. $L_{cross-entropy}$ and ~$L_{rpn\_reg}$ are the softmax loss and L1 loss for RCNN network. $L_{rpn} + L_{rcnn}$ are jointly optimized in the Faster-RCNN framework, and $L_{rcnn}$ is optimized in the Fast-RCNN framework.

In the backward stage, the gradient is computed by back-propagating the classification loss only on the ground-truth category, which is used for inverted attention generation module. With the generated Inverted Attention map, an element-wise product layer between feature maps and IA heat-maps is used for feature refinement, as
\begin{equation}
F_{new} = F .* A, 
\label{eq:refine}
\end{equation}
where $.*$ indicates element-wise multiplication. The refinement is conducted at element-level, i.e., along both the spatial and channels dimensions of the feature maps. 

After these operations, the refined features are forwarded to compute the detection loss, and then the loss is back-propagated to update the original network parameters. 
The training process of IAN is summarized in Algorithm~\ref{algo:train}. 

\begin{algorithm}
  \caption{Training Process of IAN}
  \label{algo:train}
  \begin{algorithmic}[1]
  \REQUIRE The images with ground-truth ${x_{i},y_{i}}$ ($i=1,\cdots, n$).
  \ENSURE Object detection model.
  \FOR {each iteration}
    \STATE Generating region proposal by the RPN network;
    \STATE Getting the the feature map of the region proposal by ROI pooling, as $F$ shown in Fig. \ref{fig:network};
    \STATE Computing gradient $G$ by back-warding the classification loss on the ground-truth category;
    \STATE Computing the gradient-guided attention map with Eq. \ref{eq:mask};
    \STATE Achieving spatial-wise and channel-wise inverted attention maps with Eq. \ref{eq:sa} and Eq. \ref{eq:ca};
    \STATE Refining feature map $F$ with inverted attention map with Eq. \ref{eq:refine};
    \STATE Computing RPN loss and classification loss with Eq. \ref{equ:l1} and Eq. \ref{eq:cls};
    \STATE Back-propagation.
  \ENDFOR
  
  \end{algorithmic}
\end{algorithm}

\subsection{Discussion}
The high attention regions learned by the original network represent the most common features shared by the training samples. These features are discriminative enough on the training data while may not be enough for the testing data, especially when high attention regions are corrupted by the unseen image defects.

Most top improvements on original networks were reached by discovering more discriminative features. For instance, Image based Hide-and-Seek (HaS) \cite{zhong2017random} and feature based A-Fast-RCNN \cite{wang2017fast}. HaS randomly hides patches in a training image, forcing the network to seek discriminative features on remaining patches of the images. A-Fast-RCNN finds the best patches to occlude by estimating a occlusion mask with a generation and adversary network. 

Our IA approach fuses the advantages of both approaches in the training steps of the original detector network. The new discriminative features are iteratively discovered (see Fig.~\ref{fig:IAepoch}) by inverting the original attention. IA finds discriminative features in all object parts, feature channels and even context.   This process requires no extra training epochs on hard samples and no extra network parameters to estimate occlusion mask.  

Note that, IAN is not limited to two-stage detectors like Fast-RCNN and Faster R-CNN. We also tried IAN on single-stage detectors such as Single Shot MultiBox Detector(SSD) in our experiments section 4.2.2.

\section{Experiments}

Inverted Attention Network (IAN) was evaluated on three widely used benchmarks: the PASCAL VOC2007, PASCAL VOC2012~\cite{everingham2010pascal}, and MS-COCO~\cite{lin2014microsoft} datasets. In the following section, we first introduce the experimental settings, then analyze the effect of the Inverted Attention module. Finally, we report the performance of IAN and compare it with the state-of-the-art approaches.

We used Faster-RCNN, Fast-RCNN and SSD object detectors as our baselines. Our IANs are simply constructed by adding the IA module to each baseline network. VGG16 and ResNet-101 were used as the backbone feature extractors. By default, Fast R-CNN with VGG16 were used in ablation study.
The standard Mean Average Precision (mAP) \cite{DBLP:journals/ijcv/EveringhamGWWZ10} are used as the evaluation metric. For PASCAL VOC, we report mAP scores using IoU thresholds at $0.5$. For the COCO database, we use the standard COCO AP metrics.

\subsection{Experimental Settings}


\textbf{PASCAL VOC2007:}
All models were trained on the VOC2007 trainval set and the VOC2012 trainval set, and tested on the VOC2007 test set. For Fast-RCNN, we followed the training strategies in~\cite{wang2017fast}: set the learn rate to $2e^{-3}$ for the first 6 epochs, and decay it to $2e^{-4}$ for another 2 epochs. We used batch size 2 in training, and used VGG16 as the backbone networks for all the ablation study experiments on the PASCAL VOC dataset. For Faste-RCNN, we followed the training strategies in~\cite{ren2015faster}: set the learn rate to $2e^{-3}$ for the first 6 epochs, and decay it to $2e^{-4}$ for another 2 epochs. We used the batch size 2 in training. We also report the results of ResNet101 backbone on these models. For SSD, we followed the training strategies in~\cite{liu2016ssd}: set the learn rate to $1e^{-3}$ for the first 6 epochs, and decay it to $1e^{-4}$ and $1e^{-5}$ for another 2 and 1 epochs. We use the default batch size 32 in training.

\textbf{PASCAL VOC2012:}
All models were trained on the VOC2007 trainval set, VOC2012 trainval set and VOC2007 test set, and then tested on the VOC2012 test set. For Faster-RCNN and SSD experiments, we follow the exact same training strategies as the VOC2007 training above.

\textbf{COCO:}
Following the standard COCO protocol, training and evaluation were performed on the $120k$ images in the trainval set and the $20k$ images in the test-dev set respectively. For Faster-RCNN, we set the learn rate to $1e^{-2}$ for the first 4 epochs, decay it to $1e^{-3}$ for another 4 epochs and $1e^{-4}$ for the last 2 epochs. We used batch size 16 in training, and used ResNet101 as the backbone networks. For SSD, we set the learn rate to $1e^{-3}$ for the first 6 epochs, and decay it to $1e^{-4}$ and $1e^{-5}$ for another 2 and 1 epochs. We use the default batch size 32 in training. 
\subsection{Implement Details}
In the two-stage detection framework, given the feature maps after the ROI pooling layer, we reweight the feature maps guided by inverted attention. For spatial-wise inverted attention map $H \times W$, we dropout top $33\%$ pixels with highest values. For channel-wise inverted attention map, we dropout top $80\%$ pixels with highest values.
In the single-stage detection framework, given several feature maps with different size, we follow the same strategies in above two-stage framework to reweight the feature maps. In order to prevent netwrok from overfitting to the reweighted feature maps, we only reweight $20\%$ feature maps, and leave the rest $80\%$ feature maps unchanged.

\subsection{Results on PASCAL VOC2007}

To verify the effectiveness of Inverted Attention, we first conducted ablation study on two key factors of IA, \textit{i.e.}, inversion strategies and inversion orientation. By taking the best settings in the ablation study, we then show the results compared with baselines and state-of-the-art.

\subsubsection{Ablation Study}

The following ablation study is conducted on the Fast-RCNN with the VGG16 backbone.

\begin{figure}[t]
   \centering
     \begin{subfigure}[b]{0.23\linewidth}
       \includegraphics[width=\linewidth]{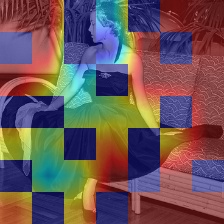}
       \caption{}
       \label{fig:invertedstrategies:Random}
     \end{subfigure}
    \begin{subfigure}[b]{0.23\linewidth}
       \includegraphics[width=\linewidth]{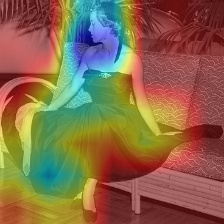}
       \caption{}
       \label{fig:invertedstrategies:Overturn}
     \end{subfigure}
     \begin{subfigure}[b]{0.23\linewidth}
       \includegraphics[width=\linewidth]{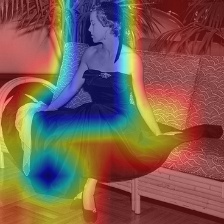}
       \caption{}
       \label{fig:invertedstrategies:hard}
     \end{subfigure}
     \begin{subfigure}[b]{0.23\linewidth}
      \includegraphics[width=\linewidth]{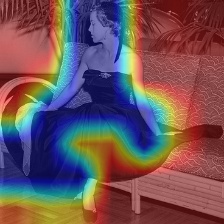}
      \caption{}
      \label{fig:invertedstrategies:soft}
     \end{subfigure}
     \caption{Visualization of four inversion strategies. From (a) to (d), it illustrates inverted attention map by random, overturn, hard-threshold, and soft-threshold, respectively.}
     \label{fig:invertedstrategies}
\end{figure}

\textbf{Inversion Strategies:} Four inversion strategies were evaluated: Random inversion, Overturn inversion, Hard-threshold inversion, and Soft-threshold inversion. Table \ref{table:is} shows that all the four inverting strategies improve the performance of the baseline. Fig.~\ref{fig:invertedstrategies} visualize the four strategies on the same on the same object. 

As shown in Fig.~\ref{fig:invertedstrategies:Random}, by randomly selecting pixels on convolutional feature maps and setting them to $0$, the mAP improves from $69.1\%$ to $70.3\%$. However, random inversion loses the contextual information, meaning that even in the same semantic part, some pixels were kept while others were discarded. As shown in Fig. \ref{fig:invertedstrategies:Overturn} Overturn inversion achieves inverted attention map by IA = 1.0 - A, which increases the weight of background and suppresses the foreground. Overturn inversion gets the mAP of 70.6\%, which is $1.5$ percentages better than baseline. To keeping the weights of background, we take two thresholding methods to suppresses all pixels in attention map which are large than the threshold. The hard-threshold is shown in Fig. \ref{fig:invertedstrategies:hard}, which takes $0.5$ as threshold. While the soft-threshold adopts a sorting algorithm and suppresses the top $33\%$ pixels. Hard-threshold achieves $70.9\%$ mAP and soft-threshold achieves $71.4\%$, which are $1.8\%$ and $2.3\%$ better than the baseline, respectively.

\begin{table}[t]
\centering 
\begin{tabular}{c | c } 
\hline 
Method   & mAP \\ [0.5ex] 
\hline \hline
Baseline (Fast-RCNN + VGG16)  & 69.1 \\
Random   & 70.3  \\ 
Overturn   & 70.6  \\ 
Hard-threshold   & 70.9  \\ 
Soft-threshold   & \textbf{71.4} \\ [0.5ex]
\hline 
\end{tabular}
\caption{Ablation study on inversion strategies.} 
\label{table:is} 
\end{table}

\begin{table}[t]
\centering 
\setlength{\tabcolsep}{2.2pt}
\begin{tabular}{c |c } 
\hline 
Method &  mAP \\ [0.5ex] 
\hline \hline
Spatial &  71.4  \\ 
Channel &  70.9\\ [0.5ex]
Spatial + Channel &  \textbf{71.6}\\ [0.5ex]
\hline 
\end{tabular}
\caption{Ablation study on inverted orientations.} 
\label{table:io} 
\end{table}

\begin{table}[t]
\centering 
\begin{tabular}{l|l|l}
\hline
                              & Method                & mAP  \\
\hline\hline
\multirow{2}{*}{ROI Feature IA}  & Fast-RCNN             & 69.1 \\
                              & Fast-RCNN + IA (ours) & \textbf{71.4} \\
\hline
\multirow{4}{*}{Full Feature IA} & Fast-RCNN             & 69.1 \\
                              & Fast-RCNN + IA (ours) & \textbf{70.2} \\
\cline{2-3}
                              & SSD300                   & 77.2 \\
                              & SSD300 + IA (ours)       & \textbf{77.8} \\ [0.5ex]
\hline 
\end{tabular}
\caption{Full feature IA v.s. ROI feature IA on VOC2007.} 
\label{table:ssd} 
\end{table}

\begin{table*}[t]
\footnotesize
\centering 
\setlength{\tabcolsep}{1.8pt}
\begin{tabular}{c c c| c| c c c c c c c c c c c c c c c c c c c c} 
\hline 
Method & Train & Backbone & mAP & aero & bike & bird & boat & bottle & bus & car & cat & chair & cow & table & dog & horse & bike & person & plant & sheep & sofa & train & TV \\ [0.5ex] 
\hline\hline 
FRCN~\cite{girshick2015fast} & 07 & VGG16 & 69.1 & \textbf{75.4} & 80.8 & 67.3 & 59.9 & 37.6 & \textbf{81.9} & 80.0 & 84.5 & 50.0 & 77.1 & 68.2 & 81.0 & 82.5 & 74.3 & 69.9 & 28.4 & 71.1 & 70.2 & 75.8 & 66.6 \\ 
ORE~\cite{zhong2017random} & 07 & VGG16 & 71.0 & 75.1 & 79.8 & 69.7 & \textbf{60.8} & 46.0 & 80.4 & 79.0 & 83.8 & 51.6 & 76.2 & 67.8 & 81.2 & \textbf{83.7} & 76.8 & 73.8 & \textbf{43.1} & 70.8 & 67.4 & 78.3 & \textbf{75.6} \\ 
Fast+ASTN~\cite{wang2017fast} & 07 & VGG16 & 71.0 & 74.4 & 81.3 & 67.6 & 57.0 & 46.6 & 81.0 & 79.3 & \textbf{86.0} & \textbf{52.9} & 75.9 & \textbf{73.7} & \textbf{82.6} & 83.2 & \textbf{77.7} & 72.7 & 37.4 & 66.3 & \textbf{71.2} & 78.2 & 74.3 \\
Fast+IA(ours) & 07 & VGG16 & \textbf{71.6} & 74.9 & \textbf{82.0} & \textbf{71.8} & 59.1 & \textbf{47.6} & 80.9 & \textbf{80.5} & 85.2 & 51.2 & \textbf{77.2} & 71.6 & 81.3 & 83.6 & 77.0 & \textbf{74.1} & 39.3 & \textbf{71.1} & 70.0 & \textbf{79.2} & 74.0 \\ 
\hline
FRCN~\cite{girshick2015fast} & 07 & ResNet101 & 71.8 &  \textbf{78.7} & 82.2 & 71.8 & 55.1 & 41.7 & 79.5 & 80.8 & 88.5 & 53.4 & 81.8 & 72.1 & 87.6 & 85.2 & 80.0 & 72.0 & 35.5 & 71.6 & 75.8 & \textbf{78.3} & 64.3\\ 
Fast+ASTN~\cite{wang2017fast} & 07 & ResNet101 & 73.6 & 75.4 & \textbf{83.8} & 75.1 & 61.3 & 44.8 & \textbf{81.9} & \textbf{81.1} & 87.9 & 57.9 & 81.2 & \textbf{72.5} & 87.6 & \textbf{85.2} & 80.3 & 74.7 & \textbf{44.3} & 72.2 & 76.7 & 76.9 & 71.4 \\
Fast+IA(ours) & 07 & ResNet101 & \textbf{74.7} & 77.3 & 81.2 & \textbf{78.1} & \textbf{62.6} & \textbf{52.5} & 77.8 & 80.0 & \textbf{88.7} & \textbf{58.6} & \textbf{81.8} & 71.4 & \textbf{87.9} & 84.2 & \textbf{81.4} & \textbf{76.6} & 44.0 & \textbf{77.1} & \textbf{79.1} & 76.9 & \textbf{77.2}\\
\hline \hline
Faster~\cite{ren2015faster} & 07 & VGG16 & 69.9 & 70.0 & \textbf{80.6} & \textbf{70.1} & \textbf{57.3} & 49.9 & 78.2 & 80.4 & 82.0 & \textbf{52.2} & 75.3 & \textbf{67.2} & 80.3 & 79.8 & 75.0 & 76.3 & 39.1 & 68.3 & 67.3 & \textbf{81.1} & 67.6 \\ 
Faster+IA(ours) & 07 & VGG16 & \textbf{71.1} & \textbf{73.4} & 78.5 & 68.3 & 54.7 & \textbf{56.1} & \textbf{81.0} & \textbf{85.5} & \textbf{84.3} & 48.4 & \textbf{77.9} & 61.7 & \textbf{80.5} & \textbf{82.6} & \textbf{75.3} & \textbf{77.5} & \textbf{47.0} & \textbf{71.7} & \textbf{68.8} & 76.0 & \textbf{72.5} \\
\hline
Faster & 07 & ResNet101 & 75.1 & 76.5 & 79.7 & 77.7 & 66.4 & 61.0 & 83.3 & 86.3 & \textbf{87.5} & 53.6 & 81.1 & 66.9 & 85.3 & \textbf{85.1} & 77.4 & \textbf{78.9} & \textbf{50.0} & 74.1 & \textbf{75.8} & 78.9 & 75.4 \\ 
Faster+IA(ours) & 07 & ResNet101 & \textbf{76.5} & \textbf{77.9} & \textbf{82.9} & \textbf{78.4} & \textbf{67.2} & \textbf{62.2} & \textbf{84.2} & \textbf{86.9} & 87.2 & \textbf{55.5} & \textbf{85.6} & \textbf{69.1} & \textbf{87.0} & 85.0 & \textbf{81.4} & 78.8 & 48.4 & \textbf{79.4} & 75.0 & \textbf{83.2} & \textbf{75.4} \\
\hline
Faster~\cite{ren2015faster} & 07+12 & VGG16 & 73.2 & 76.5 & 79.0 & 70.9 & 65.5 & 52.1 & 83.1 & 84.7 & 86.4 & 52.0 & \textbf{81.9} & 65.7 & 84.8 & 84.6 & 77.5 & 76.7 & 38.8 & 73.6 & 73.9 & 83.0 & 72.6\\ 
Faster+IA(ours) & 07+12 & VGG16 & \textbf{76.8} & \textbf{78.5} & \textbf{81.1} & \textbf{76.8} & \textbf{67.2} & \textbf{63.9} & \textbf{87.1} & \textbf{87.7} & \textbf{87.8} & \textbf{59.3} & 81.1 & \textbf{72.9} & \textbf{84.8} & \textbf{86.7} & \textbf{80.5} & \textbf{78.7} & \textbf{50.9} & \textbf{76.9} & \textbf{74.2} & \textbf{83.1} & \textbf{76.5} \\ 
\hline
Faster~\cite{ren2015faster} & 07+12 & ResNet101 & 76.4 &  79.8 & 80.7 & 76.2 & 68.3 & 55.9 & 85.1 & 85.3 & \textbf{89.8} & 56.7 & \textbf{87.8} & 69.4 & \textbf{88.3} & \textbf{88.9} & 80.9 & 78.4 & 41.7 & 78.6 & 79.8 & 85.3 & 72.0 \\ 
Faster+IA(ours) & 07+12 & ResNet101 & \textbf{81.1} & \textbf{85.3} & \textbf{86.8} & \textbf{79.7} & \textbf{74.6} & \textbf{69.4} & \textbf{88.4} & \textbf{88.7} & 88.8 & \textbf{64.8} & 87.3 & \textbf{74.7} & 87.7 & 88.6 & \textbf{85.3} & \textbf{83.5} & \textbf{53.9} & \textbf{82.7} & \textbf{81.5} & \textbf{87.8} & \textbf{80.9}\\[0.5ex]
\hline 
\end{tabular}
\caption{Object detection Average Precision (AP) tested on VOC2007.} 
\label{table:voc07} 
\end{table*}
\begin{figure*}[t]
  \centering
     \begin{subfigure}[b]{0.26\textwidth}
      \includegraphics[width=1\textwidth]{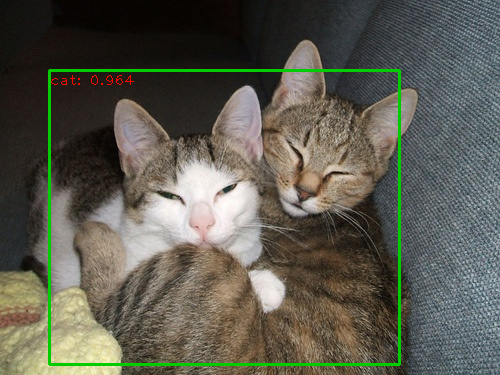}
     \end{subfigure}
    \begin{subfigure}[b]{0.29\textwidth}
      \includegraphics[width=1\textwidth]{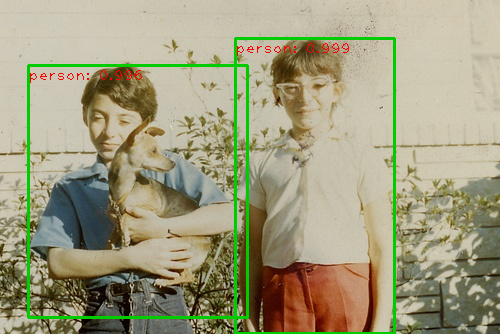}
     \end{subfigure}
     \begin{subfigure}[b]{0.13\textwidth}
      \includegraphics[width=1\textwidth]{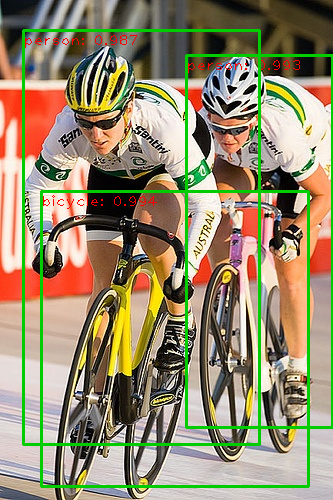}
     \end{subfigure}
     \begin{subfigure}[b]{0.26\textwidth}
      \includegraphics[width=1\textwidth]{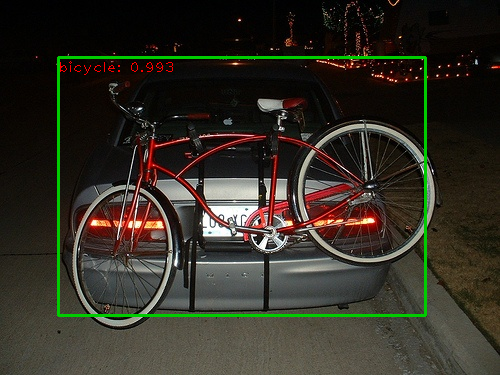}
     \end{subfigure}
      \begin{subfigure}[b]{0.26\textwidth}
      \includegraphics[width=1\textwidth]{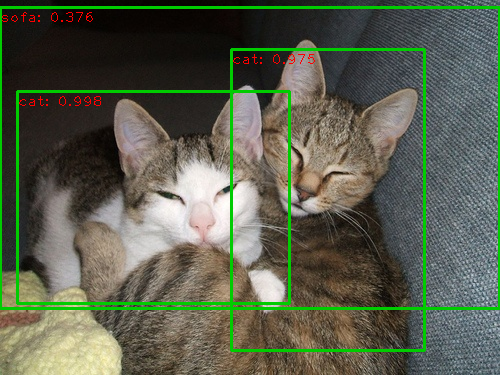}
     \end{subfigure}
    \begin{subfigure}[b]{0.29\textwidth}
      \includegraphics[width=1\textwidth]{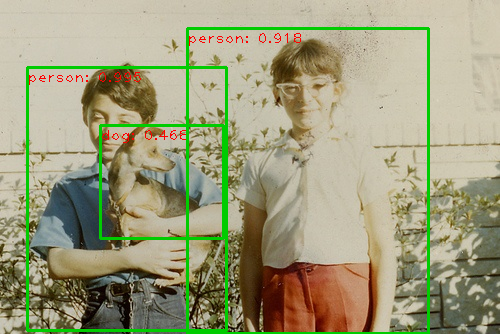}
     \end{subfigure}
     \begin{subfigure}[b]{0.13\textwidth}
      \includegraphics[width=1\textwidth]{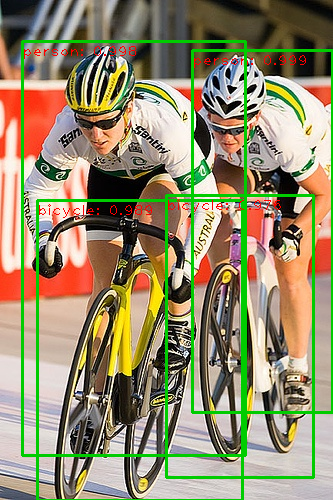}
     \end{subfigure}
     \begin{subfigure}[b]{0.26\textwidth}
      \includegraphics[width=1\textwidth]{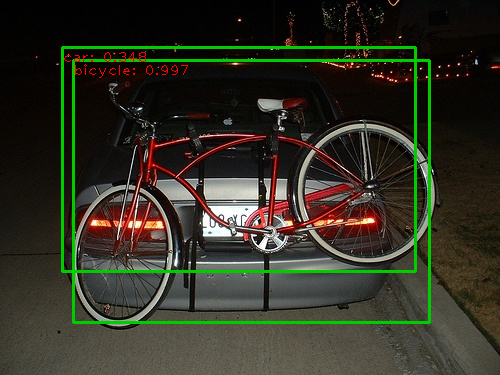}
     \end{subfigure}
     \vspace{-0.5em}
     \caption{Object detection examples in VOC2007 with ResNet101-Faster-RCNN (top) and its IAN version (bottom).} 
     \label{fig:bbox}
\end{figure*}

\textbf{Inversion Orientation:}
Using the soft-threshold inversion strategy, we further studied two inversion orientations in Table~\ref{table:io}. The spatial inversion attention conducts inversion over all channels, while the channel inversion attention conducts inversion only on a subset of channels.
Conducting spatial or channel inversion produced $71.4\%$ and $70.9\%$, respectively. Conducting both the spatial and channel inversion, the performance is further improved to $71.6\%$.



\subsubsection{Full Feature IA v.s. ROI Feature IA}

Inverted Attention is a plugin module attached to certain feature maps of neural networks. We illustrate the choices of feature maps to conduct IA on the two-stage object detection framework, e.g., Fast-RCNN and Faster R-CNN, and the single-stage object detection framework, \textit{e.g.}, Single Shot MultiBox Detector(SSD). 

In Fast-RNN, IA can be plugged on the feature maps either before, called Full Feature IA, or after ROI pooling, called ROI Feature IA. In Full Feature IA, The features of whole image is refined, while in ROI Feature IA, the features of different ROIs are refined separately. The performance comparison is shown in Table \ref{table:ssd}. IA improves the baselines no matter which strategy was used. ROI Feature IA performs better than the Full Feature IA. We believe this is because that ROI Feature IA confined the inverted attention in the vicinity of objects and is more likely to learn object-related features. Whereas for the Full Feature IA, the inverted attention covers mostly the background, making it difficult to learn object-related features. 

As there is no ROI pooling in SSD, the IA can only be plugged on the full image feature. The mAP of SSD300 was increased from $77.2\%$ to $77.8\%$.  

\subsubsection{Comparing with Baselines and State-of-the-Art }

We first present extensive performance comparison on the PASCAL VOC 2007 with Fast-RCNN (denoted as ``FRCN"), Faster-RCNN (denoted as ``Faster") and the state-of-the-art hard-sample generation approaches ORE~\cite{zhong2017random} and Fast-RCNN with ASTN~\cite{wang2017fast} (denoted as ``Fast+ASTN"). These approaches only provided results on Fast-RCNN. The results are comparied in Table \ref{table:voc07}. 

\begin{table*}[t]
\footnotesize
\centering 
\setlength{\tabcolsep}{1.6pt}
\begin{tabular}{c c c| c| c c c c c c c c c c c c c c c c c c c c} 
\hline 
Method & Train & Backbone & mAP & aero & bike & bird & boat & bottle & bus & car & cat & chair & cow & table & dog & horse & bike & person & plant & sheep & sofa & train & TV \\ [0.5ex] 
\hline\hline 
Faster~\cite{ren2015faster} & 07++12 & ResNet101 & 73.8 & 86.5 &  81.6 & 77.2&  58.0& 51.0 & 78.6 & 76.6 & 93.2 & 48.6 & 80.4 & 59.0 & \textbf{92.1} & 85.3 & 84.8 & 80.7 & 48.1 & 77.3 & 66.5 &  84.7 & 65.6 \\ 
Faster+IA(ours) & 07++12 & ResNet101 &  \textbf{79.2} & \textbf{87.7} & \textbf{86.7} & \textbf{80.3} & \textbf{68.1} & \textbf{62.1} & \textbf{81.0} & \textbf{84.7} & \textbf{93.8} & \textbf{61.8} & \textbf{84.2} & \textbf{63.1} & 92.0 & \textbf{87.4} & \textbf{86.6} & \textbf{85.8} & \textbf{61.0} & \textbf{84.6} & \textbf{72.4} & \textbf{86.5} & \textbf{73.8}\\ [0.5ex]
\hline 
SSD300~\cite{liu2016ssd} & 07++12 & VGG16 & 75.8 & \textbf{88.1} & 82.9 & 74.4 & 61.9 & 47.6 & \textbf{82.7} & 78.8 & 91.5 & 58.1 & 80.0 & 64.1 & 89.4 & \textbf{85.7} & 85.5 & 82.6 & 50.2 & 79.8 & \textbf{73.6} & \textbf{86.6} & \textbf{72.1} \\ 
SSD300+IA(ours) & 07++12 & VGG16 & \textbf{77.9} & 87.5 & \textbf{85.0} & \textbf{79.1} & \textbf{66.6} & \textbf{60.4} & 80.0 & \textbf{83.6} & \textbf{92.3} & \textbf{59.8} & \textbf{82.3} & \textbf{64.8} & \textbf{89.9} & 85.6 & \textbf{85.7} & \textbf{84.5} & \textbf{59.5} & \textbf{82.2} & 71.8 & 85.8 & 71.6 \\ [0.5ex]
\hline 
\end{tabular}
\caption{\normalsize{Object detection Average Precision (AP) tested on VOC2012.}} 
\label{table:voc12} 
\end{table*}
\begin{table*}
\centering 
\footnotesize
\begin{tabular}{c c c | c c c | c c c} 
\hline 
Method & Train & Backbone & $\mathrm{AP_{[0.5,0.95]}}$ & $\mathrm{AP_{0.5}}$ & $\mathrm{AP_{0.75}}$ & $\mathrm{AP_{s}}$ & $\mathrm{AP_{m}}$ & $\mathrm{AP_{l}}$ \\  
\hline \hline 
Faster~\cite{ren2015faster} & trainval & VGG16 & 21.9 & 42.7 & 23.0 & 6.7 &25.2 &36.4  \\ 
Faster+++~\cite{ren2015faster} & trainval35k & ResNet101 & 34.9 & 55.7 & 37.4 & \textbf{15.6} & 38.7 & 50.9 \\ 
Faster+IA(ours) & trainval35k & ResNet101 & \textbf{35.5} & \textbf{56.1} & \textbf{38.2} & 14.9 & \textbf{38.8} & \textbf{51.7}\\ 
\hline 
SSD512~\cite{ren2015faster} & trainval35k & VGG16 & 28.8& 48.5& 30.3& 10.9& 31.8& \textbf{43.5} \\ 
SSD512+IA(ours) & trainval35k & VGG16 & \textbf{29.6} & \textbf{49.8} & \textbf{31.4} & \textbf{11.9} & \textbf{32.4} & 42.8 \\ 
\hline 
\end{tabular}
\caption{Object detection Average Precision (AP) tested on COCO test-dev 2017.} 
\label{table:coco} 
\end{table*}

With the VGG16 backbone and the VOC2007 training data, IA improved Fast-RCNN from $69.1\%$ to $71.6\%$, and improves Faster-RCNN form $69.9\%$ to $71.1\%$, which are $2.5\%$ and $1.2\%$ improvement respectively. With more powerful backbone, \textit{i.e.}, ResNet101, Fast-RCNN and Faster-RCNN achieves better object detection performance than VGG16. By adding IA to them, the performance were consistently improved: for Fast-RCNN from $71.8\%$ to $71.4\%$, and for Faster-RCNN from $75.1\%$ to $76.5\%$, respectively.
Using the training data from both VOC2007 and VOC2012, the mAPs of our approach were further improved to $76.8\%$ with VGG16, and $81.1\%$ with ResNet101. 

Fig.~\ref{fig:bbox} shows some detection examples from the VOC2007 test set using ResNet101 Faster-RCNN and its IAN version. These examples illustrate that IAN improves object detection to handle images defects such as heavy occlusions, faded pictures and shadows.

\subsection{Results on PASCAL VOC2012 and COCO2017}

For the PASCAL VOC2012 and COCO2017 datasets, we used Faster-RCNN to construct Inverted Attention Network. For both of the datasets, the evaluation results were produced by the official evaluation server.

The detection performance of PASCAL VOC2012 is shown in Table~\ref{table:voc12}. For ResNet101 Faster-RCNN, IA increased mAP from $73.8\%$ of the baseline to $79.2\%$, which is $5.4\%$ improvement. The AP on 19 categories achieved consistent performance gain. For ResNet101 SSD300, IA increased mAP from $75.8\%$ of baseline to $77.9\%$, which is $2.1\%$ improvement. The AP on 14 categories achieved consistent performance gain.
This demonstrates that our IAN can discover more discriminative features for a large variety of object classes.

The detection performance of COCO2017 is shown in Table~\ref{table:coco}. We used the official evaluation metrics for COCO. Faster-RCNN baseline with VGG16 trained on the train and validation set of COCO produced $21.9\%$ for $AP_{[0.5,0.95]}$. Using the more powerful backbone ResNet101, $AP_{[0.5,0.95]}$ reached $35.5\%$ even when the training data were reduced to the train and val35k subsets of COCO. SSD512 with VGG16 produced $AP_{[0.5,0.95]}$ $28.8$. We plugged the IA module into Faster-RCNN with ResNet101 and SSD512 with VGG16 and consistently improved $AP_{[0.5,0.95]}$ $AP_{0.5}$ and $AP_{0.75}$ over the baselines. 

The COCO evaluation server also gave the detection performance on small ($AP_s$), medium($AP_m$), and large objects($AP_l$). It is interesting to note that, in Table~\ref{table:coco}, IAN tends to boost the performance of the medium and large objects for Faster-RCNN where IA was conducted on ROI features. While for SSD512, when IA was conducted on full features, IAN favors small and medium objects.

\section{Conclusion}
We present IA as a highly efficient training module to improve the object detection networks. IA computes attention using gradients of feature maps during training, and iteratively inverts attention along both spatial and channel dimension of the feature maps. The object detection network trained with IA spreads its attention to the whole objects. As a result, IA effectively improves diversity of features in training, and makes the network robust to image defects. It is very attractive to explore the best configurations of IA module on all other computer vision tasks such as image classification, instance segmentation and tracking.

{\small
\bibliographystyle{ieee}
\bibliography{egbib}
}

\end{document}